# A Chinese POS Decision Method Using Korean Translation Information.

**Son-Il Kwak**
College of Computer Science, Kim Il Sung University,
Pyongyang, DPR of Korea

**O-Chol Kown**
College of Computer Science, Kim Il Sung University,
Pyongyang, DPR of Korea

**Chang-Sin Kim**
College of Computer Science, Kim Il Sung University,
Pyongyang, DPR of Korea

**Yong-Il Pak**
College of Computer Science, Kim Il Sung University,
Pyongyang, DPR of Korea

**Gum-Chol Son**
College of Computer Science, Kim Il Sung University,
Pyongyang, DPR of Korea

**Chol-Jun Hwang**
Electronic Library, Kim Il Sung University,
Pyongyang, DPR of Korea

**Hyon-Chol Kim**
College of Computer Science, Kim Il Sung University,
Pyongyang, DPR of Korea

**Hyok-Chol Sin**
College of Computer Science, Kim Il Sung University,
Pyongyang, DPR of Korea

**Gyong-Il Hyon**
College of Computer Science, Kim Il Sung University,
Pyongyang, DPR of Korea

**Sok-Min Han**
College of Computer Science, Kim Il Sung University,
Pyongyang, DPR of Korea

**Abstract :** In this paper we propose a method that imitates a translation expert using the Korean translation information and analyse the performance.
Korean is good at tagging than Chinese, so we can use this property in Chinese POS tagging.

**Introduction**

Previous POS(Part Of Speech) tagging methods of Chinese can be largely classified into 2.

One is a method using POS tagging rules that is extracted by Chinese experts, and the other is to use a statistical model, it usually needs huge Chinese corpus.

In case of use the first method, it is very difficult to extract rules and these rules can't represent various language examples.

So this method was used in the early years of POS tagging and has been used together with other methods.

Other method, corpus-based statistical methods have many advantages, but it needs large amount of POS corpus.

To solve these problems, some researchers are studying about a method that uses raw text data extracted from internet[2, 5, 6].

In this paper we propose a POS decision method imitate Chinese translator whose native language is Korean and POS decision rule extracting method using Chinese –Korean bilingual corpus.

**1. Availability of Korean translation information in Chinese POS decision**

Machine translation system is a typical expert system, so the closer translation method to translation expert, the higher accuracy of the translation system.

Translation expert whose mother tongue is Korean generate Korean words from Chinese words when he translates Chinese, and verifies their translation and finally complete the translation.

For example, consider following sentence.

精密的观察是科学研究的基础(세밀한 관찰은 과학연구의 기초이다.)

When the expert translates this sentence, expert generate Chinese words '科学研究' from 《과학연구》, and he knows these words are right words and 《과학은 연구한다》 are not right through his experience.

Classical method is to divide the sentence as 精密/的/观察/是/科学/研究/的/基础, and decide the POS of multi-POS word'研究'using statistical information.

This method largely depends on the amount of corpus and their fields.

We can use Korean POS tagging system for this problem.

Using the POS tagged words of 《세밀한 관찰은 과학연구의 기초이다.》, we can realize the noun combination of 《과학연구》 and it tells us the POS of Chinese word '研究'is noun.

In conclusion using large amount of Korean POS tagged corpus, we can improve the performance of Chinese POS tagging system.

**2. Acquisition method of POS decision rule from Chinese-Korean bilingual corpus.**

Above-mentioned method can be applied into noun - noun conjunction, but it can not be applied into verb decision and so on.

Generally we use following rules for the POS decision of Chinese multi POS words.

```
<ruleset name="posdecpos" kind="vn">
<rule cond="any+word(了)">
            setpos(0,v)
</rule>
```

Acquisition of these rules generally needs many efforts and it is difficult to decide the confidence of rule.

We can use Chinese-Korean bilingual corpus to extract statistical rule.

Consider the acquisition method through a example.

Chinese sentence: 我的朋友学习中国语

Translation: 나의 동무는 중국어를 배운다

For above example, do the word dividing for Chinese and do the POS tagging for Korean.

我/的/朋友/学习/中国语

나/N 의/T 동무/N 는/T 중국어/N 를/T 배우다/V ㄴ다/T

In this sentence, morpheme analysis of 《배운다》 is 《배우다/동사 + ㄴ다/맺음토》 and verb translation of 学习 is 《배우다》, so 学习 is used in verb.

Chinese sentence $Cs$ and Korean sentence $Ks$ can be represented as follows.

$Cs : c_1 c_2 \cdots c_n$

$Ks : k_1 k_2 \cdots k_m$

where, $n, m$ are the number of Chinese words and Korean words, $c_i, k_i$ are the $i$ th word of Chinese sentence and Korean sentence.

Generally Chinese word $c_i$ has several POS.

First, do the morpheme analysis of Chinese sentence $Cs$ and Korean sentence $Ks$.

Let the result be $c_1 / c_2 / \cdots / c_i / \cdots c_n$, $k_1 / k_2 / \cdots / k_i / \cdots k_m$.

The algorithm decide the POS of $c_i$ as follows.

Using this method with Chinese POS tagging system, we can train the context of Chinese POS without POS tagged corpus.

**Verb decision algorithm**

1: Initialization

  $i = 1; j = 1$

2: $if\ HasVerb(c_i)\ then\ goto\ 3\ else\ 4$

3: $buf = Verb(c_i)$

  $for(j = 1 : m)$

  {

   $if\ (k_j == buf)$

    { $Setverb(c_i)$

     $\text{Re}cordcontext;$ }

  }

4: $i++;\ goto\ 2$

Where,

$HasVerb(c)$ : The function that find whether $c$ has verb or not

$Verb(c)$ : The function that return Korean verb translation of $c$

$Setverb(c)$ : The function that set the Chinese word $c$ to be verb

$\text{Re}cordcontext$ : The context recording function

Recoding context means the process that set the unknown properties of the POS deciding rule.

The POS deciding rule was represented as follows.

 <ruleset name="posdecpos" kind="*Deciding type*">

 <rule cond="context">

   setpos( *multi POS position*, *POS information*)

 </rule>

Unknown properties are *Deciding type, multi POS position, POS information.*

For all sentences of training data, we apply the algorithm described above and find unknown properties and record it, then can extract POS deciding rules and certainty factor.

## 3. Chinese POS tagging method

We constitute the Chinese POS tagging system with above mentioned methods.

First, we extracted noun-noun conjunction from Korean POS tagged corpus and saved it in dictionary.

And then acquired the POS tagging rule using Chinese-Korean bilingual corpus.

Finally we propose the POS tagging method as follows.

1. Do the word dividing of Chinese sentence $Cs$ and denote it be $Cs: c_1 c_2 \cdots c_n$

2. For single POS words, decides the POS of the word..

3. For the word $c_i$ that it has several POS and it includes noun, find the noun conjunction $c_{i-1} c_i$, $c_i c_{i+1}$ in noun conjunction dictionary, and if it success, tag the word to be noun.

4. Do the POS tagging using the POS tagging rule.

5. For the words untagged above, tag using statistical model as follows.[1]

$$POS_i = \arg\max_{t_i} [\frac{\lambda_2 p(t_i | T_{i+1}) p(t_i | w_i)}{p(t_i)} + \frac{\lambda_1 p(t_i | t_{i-1}) p(t_i | w_i)}{p(t_i)}]$$

where,
$\lambda_1 = 0.772$, $\lambda_2 = 0.22$

**4. Training using Korean inforamation**

To extract Korean noun conjunction information we use 200MB Korean POS tagged corpus in various field.

Korean POS tagged corpus is as follows.

상품/NNGC++의/TCP 가치/NNGC++는/TA 상품/NNGC++생산/NNGV++과/TCJ 교환/NNG++의/TCP 존재/NNGC++와/TCJ 관련되/PVG++ㄴ/TDP 경제/NNGC++범주/NNG++이/TEP++다/TFK ./NNPX 상품/NNGC++생산/NNGV++과/TCJ 교환/NNG++을/TCO 떠나/PVG++아/TJA++서/TJ++는/TA 가치/NNGC++문제/NNGC++에/TCD 대하/PVG++여/TJA 론하/PVG++ㄹ/TDF++수/NNDIP 없/PAS++다/TFK ./NNPX

We extracted 200 thousand noun conjunction pairs automatically and saved in dictionary.

Table 1 shows the examples of Korean noun conjunctions and corresponding Chinese words.

Table 1. Example of noun conjunction

| N | Korean noun conjunction | Chinese word conjunction |
|---|---|---|
| 1 | 조종/기술 | 控制/技术 |
| 2 | 국방/건설 | 国防/建设 |
| 3 | 교환/기술 | 交换/技术 |
| 4 | 유한/확률 | 有限/概率 |
| 5 | 열/전도 | 热/传导 |
| … | … | … |

And then we trains the context in 20 thousand sentences Chinese-Korean bilingual corpus and extract information of 30 thousand Chinese verb words.

After correction we can acquire following POS decision rule.

```
<ruleset name="posdecpos" kind="vn">
    <rule cond="any+word(、)+spos(n)">
        setpos(0,n)
    </rule>
    <rule cond="word(所)+any" main="1">
        setpos(1,v)
    </rule>
```

```
<rule cond="any+word(是)">
        setpos(0,n)
</rule>
```

**5. Experiment and evaluation.**

We use test data of the 1th Chinese - Korean machine translation system tournament , which was conducted by D.P.R.K and 5 teams participate in this tournament. This test data consists of 100 Chinese sentences.

We also use the previous method[1] to establish a baseline and evaluate our result with best reported results in the literature.

Table 2 shows errors of our POS tagging system.

Table 2.　Errors of our POS tagging system

| Total number of words | Number of sentences | Word dividing errors | Pos tagging errors | Unknown words |
|---|---|---|---|---|
| １６９１ | １００ | １９ | ２５ | ３ |

And we find that noun-verb errors are 24% and verb-noun errors are 36% in the whole error.

This result says that V-N errors are main part of Chinese POS tagging error and it must be solved in the future.

Final result is tabulated in table 3.

Table 3. Final result of experiment

| Previous method | | Our method |
|---|---|---|
| Previous method 1 | Previous method 2 | |
| ９０％ | ９４％ | 98% |

Experimental results show that this effort is rewarding, and the tagging accuracy is significantly improved.

**Conclusion**

In this paper we imitate the process that Chinese translation expert whose mother tongue is Korean generate Korean and evaluate the accuracy of the system.

But we only extract noun conjunction information from Korean and can not consider verb information and only consider words that it has noun and verb when we use bilingual corpus.

In the future we must study about the POS tagging method of Chinese verb using Korean toe information and the POS decision rule.